\documentclass[conference]{IEEEtran}
\IEEEoverridecommandlockouts
\usepackage{cite}
\usepackage{amsmath,amssymb,amsfonts}
\usepackage{algorithmic}
\usepackage{graphicx}
\usepackage{multicol}
\usepackage{lipsum}
\usepackage{textcomp}
\usepackage{xcolor}
\usepackage{subcaption} 
\usepackage{multirow}
\usepackage{caption}
\usepackage{array}
\usepackage{makecell}
\usepackage{diagbox}
\usepackage{amsmath}
\usepackage{booktabs}

\usepackage{caption}
\usepackage{subcaption}

\newcommand{\CP}[1]{\ignorespaces}
\newcommand{\ie}{\textit{i.e.}}
\newcommand{\eg}{\textit{e.g.}}

\usepackage{url} 

\begin{document}


\title{Impact of 3D LiDAR Resolution in Graph-based SLAM Approaches: A Comparative Study
}

\author{\IEEEauthorblockN{J. Jorge$^{1}$, T. Barros$^{1}$, C. Premebida$^{1}$, M. Aleksandrov$^{2}$, D. Goehring$^{3}$, U.J. Nunes$^{1}$}
\IEEEauthorblockA{$^{1}$ University of Coimbra, Institute of Systems and Robotics, DEEC, Portugal\\
\{joao.jorge,~tiagobarros,~cpremebida,~urbano\}@isr.uc.pt}
$^{2}$ Data Science Institute, Universität Greifswald, Germany\\
martin.aleksandrov@uni-greifswald.de\\
$^{3}$Institute of Computer Science, Freie Universität Berlin, Germany\\
daniel.goehring@fu-berlin.de \\
}



\maketitle


\begin{abstract}
Simultaneous Localization and Mapping (SLAM) is a key component of autonomous systems operating in environments that require a consistent map for reliable localization. SLAM has been a widely studied topic for decades with most of the solutions being camera or LiDAR based. Early LiDAR-based approaches primarily relied on 2D data, whereas more recent frameworks use 3D data. In this work, we survey recent 3D LiDAR-based Graph-SLAM methods in urban environments, aiming to compare their strengths, weaknesses, and limitations. Additionally, we evaluate their robustness regarding the LiDAR resolution namely 64 $vs$ 128 channels. Regarding SLAM methods, we evaluate SC-LeGO-LOAM, SC-LIO-SAM, Cartographer, and HDL-Graph on real-world urban environments using the KITTI odometry dataset (a LiDAR with 64-channels only) and a new dataset (AUTONOMOS-LABS). The latter dataset, collected using instrumented vehicles driving in Berlin suburban area, comprises both 64 and 128 LiDARs. The experimental results are reported in terms of quantitative `metrics' and complemented by qualitative maps. 
\end{abstract}



\section{Introduction}
\label{sec:introduction}

Simultaneous Localization and Mapping (SLAM) is a critical component of autonomous systems operating in environments that require a metric map for reliable localization. SLAM has been extensively studied for several decades, with early LiDAR-based approaches predominantly relying on 2D data, while recent advancements utilize 3D data for improved accuracy and robustness.

Intelligent and autonomous vehicles (AVs), including autonomous robots, have become increasingly integrated into our daily lives and are deployed in various applications, such as transportation, delivery services, agriculture, surveillance, and industrial automation. Historically, mobile robots were typically limited to performing repetitive or predetermined tasks. However, advancements in technology have significantly altered this view. Developments in artificial intelligence (AI), recent sensor technologies (\eg, 3D LiDAR), and increased computational resources have enabled robots and AVs to navigate autonomously in real-world environments. Additionally, progress in SLAM and localization systems has further enhanced their reliability.

In SLAM, a robot or vehicle equipped with onboard sensors estimates its current position while simultaneously constructing a map of the surrounding environment~\cite{cadena2016past}. The pose (position and orientation) of the vehicle provides a comprehensive description of its current state, while the map represents a selective depiction of relevant elements, such as the locations of obstacles and landmarks, that characterize the robot's operating environment. SLAM techniques can generally be categorized into three main types: visual-based~\cite{mur2015orb,wang2017stereo}, LiDAR-based~\cite{grisetti2007improved, konolige2010efficient, kohlbrecher2011flexible}, and visual-LiDAR fusion~\cite{liang2016visual, 7139486}. Early LiDAR-based approaches primarily utilized 2D data, whereas more recent methods leverage 3D data for enhanced performance.

\begin{figure}[t]
    \centering
     \includegraphics[width=\columnwidth]{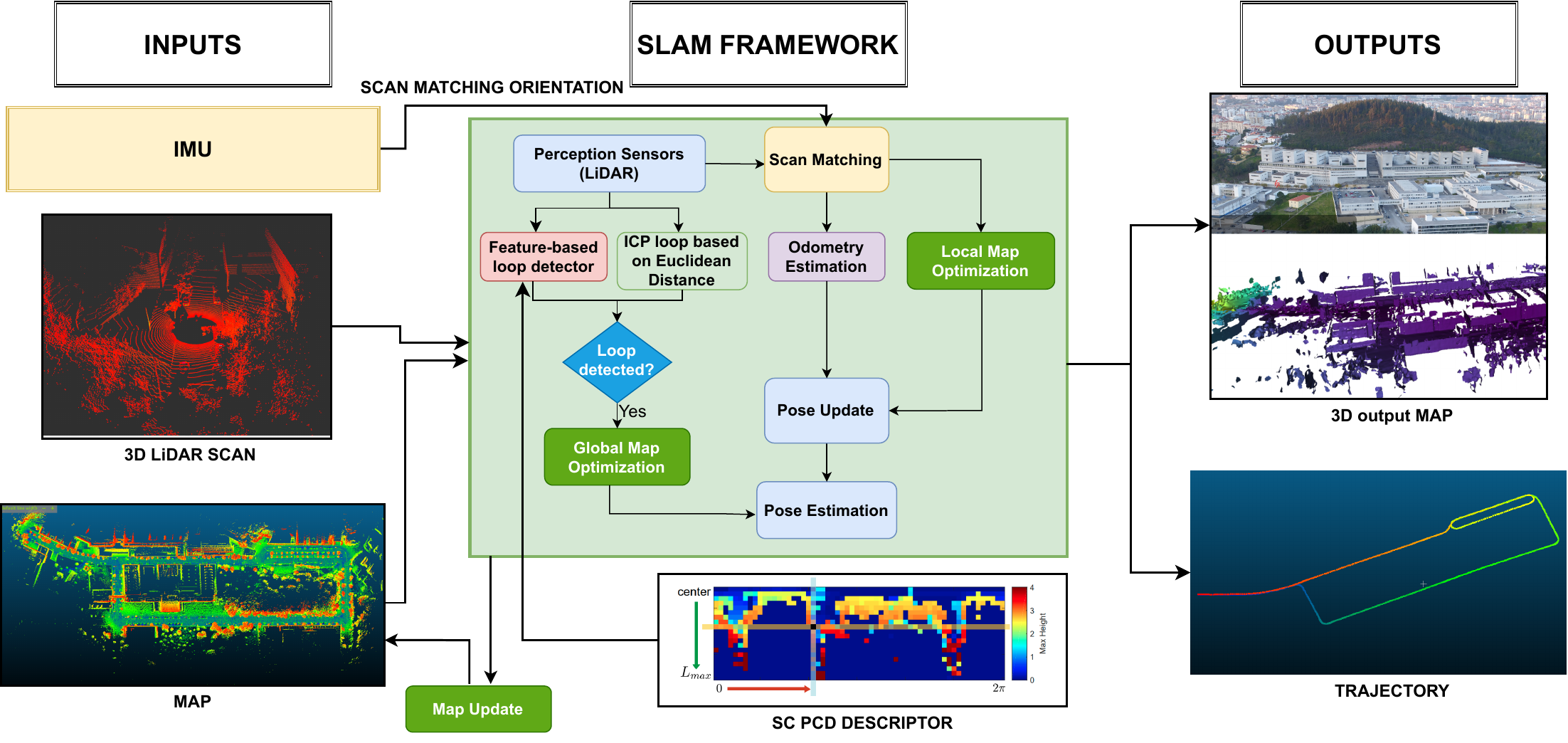}
    \caption{Example of a representative LiDAR Graph-based SLAM framework.}
    \label{fig:SCLIO-SAMframeworkpipeline}
\end{figure}

In this work, we provide a survey of recent 3D LiDAR-based Graph-SLAM methods in urban environments, with the objective of comparing their strengths, weaknesses, and limitations under varying scan resolutions. Figure~\ref{fig:SCLIO-SAMframeworkpipeline} outlines the key modules of graph-based SLAM approaches.

Specifically, we evaluate SC-LeGO-LOAM~\cite{sc-lego-loam}, SC-LIO-SAM~\cite{sc-lio-sam}, Cartographer~\cite{hess2016real}, and HDL-Graph SLAM~\cite{koide2019portable}. To assess these approaches under real-world conditions, we report experimental results using the KITTI odometry dataset and the AUTONOMOS-LABS dataset. The latter, a recent dataset collected in the suburban areas of Berlin, comprises 3D LiDAR data captured with 64- and 128-beam sensors, highlighting the importance of evaluating different sensor resolutions to understand their impact on SLAM performance.

\begin{table*}[t]
    \centering
    \caption{Default SLAM Parameters and Algorithms.}
    \label{tab:slam_params}
    {\renewcommand{\arraystretch}{1.3}
    \begin{tabular}{l|ccccc}
        \noalign{\hrule height 1pt}\hline	
        Parameter & \textbf{Cartographer} &\textbf{SC-LIO SAM} & \textbf{SC-LeGO LOAM} & \textbf{HDL-GRAPH}  \\
        \hline
        Pointcloud Registration & scan-to-submap  & edge and corner registration  & edge and corner registration &    point-to-plane \\
        Pointcloud Downsample  &  occupancy grid 0.05  & -  & -  &  0.1    \\
        Surf/Corner Feature Voxel.  &  -/-    & 0.1/1 & 0.1/0.1 & -/- \\
        Surf/Corner Map Voxel.  &   -/-  & 0.4/0.2 & 0.4/0.2 & -/- \\
        Map voxelization   &  2D map   &  -/- & -/-  & 0.05  \\
        ICP Fitness Score   &  0.45   &  0.3\  &  0.3  & 2.5 \\
        
        Loop Closure Algorithm   &   RS (scan-submap)    & SC+RS & SC+RS  & RS \\
        Loop Closure Rate   &  on keyframe update  &  1 hz    &   1 hz  & on keyframe update \\
        Graph Optimization Algorithm    &  Ceres & iSAM2  & iSAM2  & g2o\\
        Graph Optimization Rate &  every 10s & every frame   & every frame  & every 3s\\
        \noalign{\hrule height 1pt}\hline	
    \end{tabular}
    }
\end{table*}

\section{Related Work}
\label{related_work}

Simultaneous Localization and Mapping (SLAM) algorithms can be categorized into three main approaches: filter-based, graph-based, and deep learning-based SLAM. Filter-based SLAM approaches treat the SLAM problem as state estimation using probabilistic filters, such as the Extended Kalman Filter (EKF)~\cite{welch1995introduction}, Unscented Kalman Filter (UKF)~\cite{julier1997new,wan2000unscented}, and Particle Filters~\cite{NIPS2000_f5c3dd75}. Graph-based SLAM methods~\cite{zhang2014loam,sc-lio-sam,shan2020lio,sc-lego-loam,shan2018lego,koide2024glim} represent pose estimation as a pose-graph optimization problem~\cite{lu1997globally}. On the other hand, deep learning-based SLAM leverages neural networks to directly learn representations of the environment from sensor data, allowing for end-to-end training approaches to learn complex mappings between sensor measurements and the robot's pose or map~\cite{yue2020lidar}.

In this work, we focus on graph-based LiDAR-SLAM approaches, specifically studying the impact of sensor resolution on pose and map estimation. Graph-based SLAM typically involves two primary modules: a front-end, which estimates the pose and map using scan odometry, and a back-end, where the pose and map are refined using graph-based global optimization with established libraries such as Ceres~\cite{ceres-solver}, g2o~\cite{5979949}, and GTSAM~\cite{dellaert2012factor}.

In the front-end module, a critical component is LiDAR point-cloud registration, which computes a rigid transformation by minimizing alignment errors between pairs of point clouds~\cite{babin2021large}. This can be achieved using iterative methods such as Normal Distributions Transform (NDT)\cite{biber2003normal}, Random Sample Consensus (RANSAC)\cite{fischler1981random}, or the Iterative Closest Point (ICP) algorithm and its variations~\cite{besl1992method,bouaziz2013sparse,segal2009generalized}. 
ICP-based registration, the most popular among these methods, can be formulated as a point-to-point, point-to-plane, or point-to-normal association problem~\cite{pomerleau2015review}. Point-to-point ICP minimizes the distance between corresponding points, point-to-plane ICP minimizes the distance between a point and a plane, and point-to-normal ICP considers the alignment of surface normals, which can improve convergence in certain scenarios. 
However, performing registration on full point clouds is computationally intensive. To address this, recent SLAM approaches use feature extraction to identify robust key points, reducing the number of points involved in the registration process. For instance, state-of-the-art (SOTA) methods such as LOAM~\cite{zhang2014loam}, LeGO-LOAM~\cite{shan2018lego}, and LIO-SAM~\cite{shan2020lio} extract edge and planar features from the raw input scans.

In the back-end module, an essential aspect of graph-based SLAM is loop closure detection (LCD). LCD identifies revisited regions, adding constraints to the pose-graph and improving the consistency of the map. LCD is crucial for reducing drift in SLAM systems, as it helps correct cumulative errors by leveraging previously visited locations to refine the overall map and pose estimates. This is also referred to as place recognition in some literature~\cite{barros2021place}, which involves identifying previously visited locations based on visual, structural, or semantic features derived from the surrounding environment. By comparing current sensory data with past observations, the system can identify matching features or similarities. When a loop closure is detected, the system uses this information to correct and update the global map, thus enhancing the overall accuracy of the SLAM solution.
\section{Graph SLAM Frameworks and Datasets}
\label{comparison}

As mentioned earlier, this work focuses exclusively on graph-based SLAM approaches. Among several SOTA methods, we selected Cartographer, SC-LIO SAM, SC-LeGO LOAM, and HDL-Graph SLAM due to their widespread adoption in different applications. These methods were chosen for their unique features: Cartographer's efficiency in both 2D and 3D mapping, SC-LIO SAM's integration of LiDAR and IMU data for improved accuracy, SC-LeGO LOAM's real-time feature extraction, and HDL-Graph SLAM's suitability for high-resolution LiDAR data. These methods follow a common structure consisting of four key components: initial scan alignment, subsequent pose optimization, detection of previously visited locations, and map construction. Initial scan alignment involves aligning consecutive LiDAR scans to establish an initial estimate of the vehicle's movement. Subsequent pose optimization refines this estimate by minimizing alignment errors and ensuring consistency across the trajectory. Detection of previously visited locations, also known as loop closure, helps reduce drift by identifying when the vehicle returns to a known location. Finally, map construction uses the refined pose estimates to create an accurate representation of the environment. An example of a graph-based SLAM pipeline is illustrated in Fig.~\ref{fig:SCLIO-SAMframeworkpipeline}.

To evaluate the algorithms, we used the KITTI VISION benchmark~\cite{geiger2012we} and the AUTONOMOS-LABS dataset. The KITTI dataset comprises point clouds captured in various urban environments using an onboard Velodyne HDL-64E S2 and an IMU (OXTS RT 3003). The AUTONOMOS-LABS dataset was collected using a Velodyne HDL-64 (sequences 00 and 22) and a VLS-128 LiDAR (sequences 25 and 55). 

The results obtained from the AUTONOMOS-LABS dataset sequences were generated without the use of an IMU due to poor data quality. This limitation likely affected the accuracy of the pose estimates, as IMU data is typically used to provide additional information about the vehicle's orientation and motion, helping to reduce drift and improve overall SLAM performance. Except for Cartographer, we tested the algorithms using LiDAR data only (\ie, without IMU data) to study the impact of different LiDAR resolutions on SLAM. Table \ref{tab:LiDAR_64_128_Comparison} summarizes the main differences between the 64-beam and 128-beam LiDAR sensors used in the AUTONOMOS-LABS dataset.

\begin{table}[t]
    \centering
    \caption{VLS-128 and HDL-64E\_S2 specifications from the AUTONOMOS-LABS LiDARs.} 
    {\renewcommand{\arraystretch}{1.2}
    \begin{tabular}{l|cc}
        \noalign{\hrule height 0.5pt}\hline	
         LiDAR Sensor    & HDL-64E\_S2 & VLS-128 \\  \hline
        Horizontal FoV   &  360º      &  360º  \\ 
        Vertical FoV     &   +2º to -24.9º &   +15º to -25º \\ 
        Range &    120m & 300m  \\ 
        Points per Scan &    Up to 1.3M   & Up to 700m \\ 
        \noalign{\hrule height 0.5pt}\hline	    
    \end{tabular}
    }
    \label{tab:LiDAR_64_128_Comparison}
\end{table}
\section{Experimental Evaluation}
\label{sec:results}

In this section, we outline the experiments and discuss the empirical results. The experiments were conducted on the KITTI Odometry and AUTONOMOS-LABS datasets, from which only sequences containing revisited locations were selected. 

\subsection{Performance Metrics}
\label{sec:Materials}
A comprehensive evaluation was conducted based on a range of criteria, including processing time, Absolute Trajectory Error (ATE), and Relative Trajectory Error (RTE). ATE measures the overall difference between the estimated trajectory and the ground truth.
Hence, given a sequence of estimated positions $\{p_i| p_i \in \mathbb{R}^3 \}$ and the corresponding ground truth $\{g_i| g_i \in \mathbb{R}^3\}$ with $i=1,2,...,n$ samples, and assuming that both trajectories are aligned, the $\text{ATE}_{RMSE}$ is computed  as follows:

\begin{align}
    \text{ATE}_{RMSE} &= \sqrt{\left (\frac{1}{n}\sum_{i=1}^{n} || p_i - g_i || ^2 \right)}.
\end{align}

On the other hand, RTE quantifies the local drift by comparing relative positions between consecutive points in the estimated and true trajectories over a fixed time interval $\delta$. Given $P \in SE(3)$ as the pose estimation and $G\in SE(3)$ as the ground truth pose, the relative pose error $E_{R_i} \in SE(3)$ at timestamp $i$ is computed as follows:

\begin{equation}
    E_{R_i} = (G^{-1}G_{i+\delta})\cdot (P^{-1}P_{i+\delta}).
\end{equation}

For a sequence of $n$ poses, we obtain $m = n - \delta$ relative pose errors, the $\delta$ parameter was set to 10. The RTE is usually divided in two components: rotation and translation. As for the ATE, the RTE is computed based on RMSE as follows:

\begin{align}
    RTE_{trans}^{i,\delta} &= \sqrt{\left(\frac{1}{m}\sum_{i=1}^{m} || \text{trans}(E_{R_i}) || ^2\right)}, \\
    RTE_{rot}^{i,\delta} &= \sqrt{\left(\frac{1}{m}\sum_{i=1}^{m} || \angle(\text{rot}(E_{R_i} )) ||^2\right)},
\end{align}

\noindent where $\text{trans}(\cdot)$ and $\text{rot}(\cdot)$ extract the translation and rotation components respectively from $E_{R_i}$.

To evaluate the processing time in \textit{ms} of each algorithm, we calculate the average time of the main threads and took the highest value, since they are processed in parallel. 

\subsection{Implementation Details}
The individual methods were implemented using their open-source code and default parameters, with one exception: for Cartographer, we configured the Ceres\footnote{http://ceres-solver.org/} scan matcher due to poor rotation results when using the default scan matcher on the KITTI dataset. The Ceres scan matcher is an optimization-based scan matching library that provides better accuracy in handling complex rotations and non-linearities, which is why it was chosen to replace the default scan matcher.

All experiments were conducted on a machine with Ubuntu 20.04 LTS and ROS Noetic, with a ROS bag rate of 0.01 Hz, to ensure that every frame was processed, and the algorithms were configured according to their default parameters, as shown in Table \ref{tab:slam_params}. 

\subsection{Results}
The empirical results are presented in Table~\ref{tab:KITTI_ATE}, Table~\ref{tab:KITTI_ATE}, and Table~\ref{tab:fuberlin_results}. 

\begin{table*}[t]
    \centering
    \captionsetup{position=top, justification=centering} 
    \caption{ATE and RTE results on the KITTI Odometry dataset are reported in [deg]/[m].} 
    {\renewcommand{\arraystretch}{1.2}
    \begin{tabular}{c|ccccc|cccccc}
         \noalign{\hrule height 1pt}\hline	
          &  \multicolumn{5}{c|}{ATE} &  \multicolumn{5}{c}{RTE} \\ \hline 
         & 00 & 05 & 07&08 & Avg. & 00& 05& 07 & 08& Avg.\\  \hline
        Cartographer   &    -/-    & 2.45/\textbf{4.43} & 1.86/2.98 & \textbf{2.08}/\textbf{16.42} & 2.13/\textbf{7.94}  &  -/-            &              0.31/0.29               &           0.53/0.24         & 0.46/0.45 & 0.43/0.32 \\ 
        HDL-Graph  & 3.27/15.72 &      3.80/14.03   & \textbf{0.95}/\textbf{2.00} & 5.83/38.57 & 3.46/17.58 & \textbf{0.67}/\textbf{0.23} &     \textbf{0.23}/\textbf{0.10}      & 0.24/\textbf{0.15} & \textbf{0.38}/\textbf{0.37} & 0.38/\textbf{0.21} \\ 
        SC-LL & \textbf{3.06}/\textbf{9.48}   & 3.56/9.96 & 2.07/2.54  &  4.97/30.1 & 3.41/13.02 &           1.17/1.11         &               0.56/0.54              &           0.72/0.26         & 0.98/0.42 & 0.85/0.58 \\ 
        SC-LIO &        -/-    & \textbf{2.17}/7.30 & 1.24/2.22 & 2.11/17.71 & \textbf{1.84}/9.07  &              -/-            &               0.27/0.28              &           \textbf{0.22}/0.30         & 0.39/0.72 & \textbf{0.29}/0.43\\  \noalign{\hrule height 1pt}\hline	
    \end{tabular}
    }
    \label{tab:KITTI_ATE}
\end{table*}

\begin{figure*}[t]
        \centering
        \begin{subfigure}[b]{0.24\textwidth}
            \includegraphics[width=\textwidth,trim={0.5cm 1.5cm 0cm 0.8cm},clip]{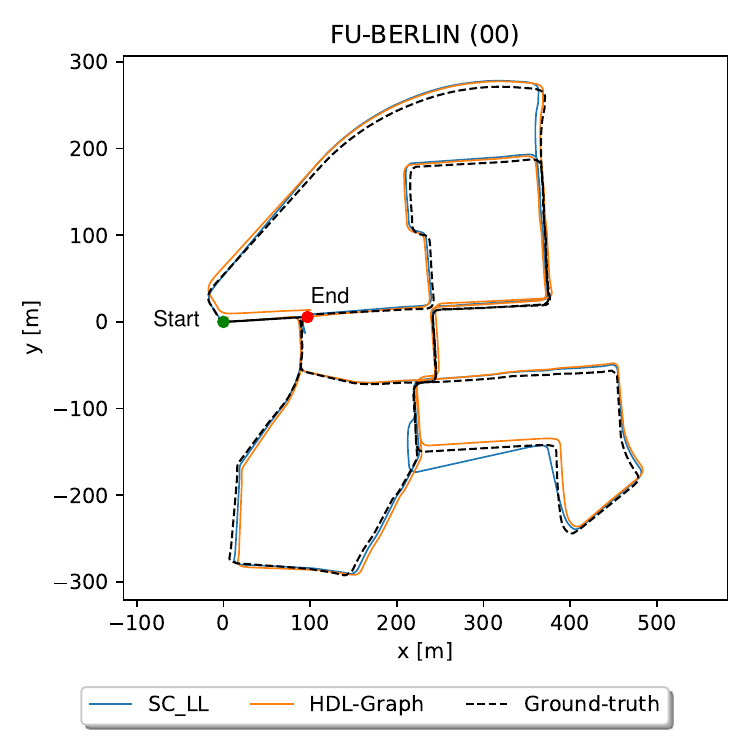}
        \end{subfigure}
        \begin{subfigure}[b]{0.74\textwidth}
            \includegraphics[width=\textwidth,trim={0.5cm 1.5cm 0.2cm 0.8cm},clip]{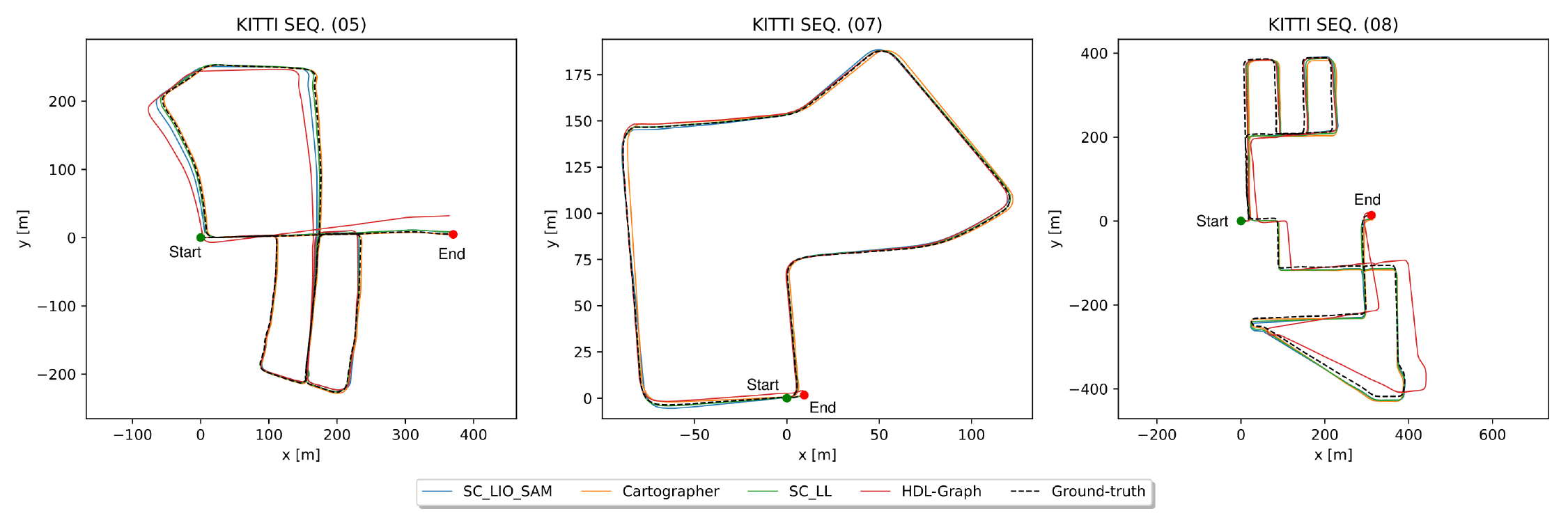}
        \end{subfigure}
        ~
        \begin{subfigure}[b]{0.5\textwidth}
            \includegraphics[width=\textwidth,trim={0cm 0cm 0cm 0cm},clip]{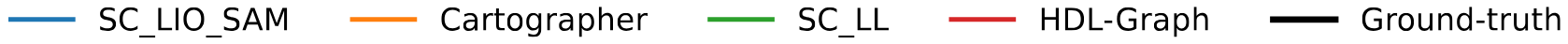}
        \end{subfigure}
        \caption{Estimated paths on the KITTI Odometry dataset.}
        \label{fig:kitti_results}
\end{figure*}

\subsubsection{KITTI Dataset}

The feature-based algorithms, SC-LIO SAM and SC-LeGO-LOAM (SC-LL), performed consistently throughout both dataset, with SC-LIO SAM showing a slight advantage. The trade-off between computational cost and performance makes both methods viable for practical use. Cartographer, while achieving reasonable results, is limited by its inability to produce a 3D map and its reliance on high-frequency IMU data. HDL-Graph SLAM, the only method using raw point cloud scan matching without feature extraction, achieved the most accurate scan matching (\ie, lower RPE) across all trajectories. However, the absence of a more robust loop closure algorithm led to less favorable overall results.

In terms of loop closure detection, HDL-Graph SLAM did not deviate significantly from the others but detected fewer correct loops compared to SC-LeGO LOAM. Once a loop was identified, SC-LeGO LOAM was able to align the trajectory accurately with the ground truth by immediately correcting its path. This can be observed at the end of KITTI sequence 00, where HDL-Graph SLAM failed to complete the loop. In KITTI sequence 07, which was the shortest trajectory, HDL-Graph SLAM achieved the best ATE and RTE but was still unable to identify a revisited location at the end, leaving the loop unclosed. In contrast, all other algorithms successfully completed the loop. For KITTI sequences 05 and 08, Cartographer outperformed both SC-LeGO LOAM and SC-LIO SAM (as shown in Fig.~\ref{fig:kitti_results}), producing reasonable trajectory estimates. The main limitation of Cartographer was its inability to generate a 3D map, despite estimating a 6-DoF (Degrees of Freedom) pose.

Regarding scan matching performance, SC-LIO SAM and Cartographer exhibited comparable performance. The superior rotational performance of SC-LIO SAM can be attributed to its use of IMU data as a front-end component, alongside LiDAR odometry. The precise orientation data from the IMU allowed SC-LIO SAM to outperform the other methods. HDL-Graph SLAM was unable to close the loop and prevent drift from the ground truth. In contrast, the remaining three algorithms successfully identified loops and corrected their trajectories accordingly.

\begin{table*}[t]
    \centering
    \captionsetup{position=top, justification=centering} 
    \caption{ATE and RTE results on the AUTONOMOS-LABS dataset for each 10 frames are reported in [deg]/[m].} 
    {\renewcommand{\arraystretch}{1.2}
    \begin{tabular}{l|cc|cc|c|c c | c c| c }
    \noalign{\hrule height 1pt}\hline	
         &  \multicolumn{5}{c|}{ATE} &  \multicolumn{5}{c}{RTE} \\ \hline 
         LiDAR Beams    & \multicolumn{2}{c|}{64} & \multicolumn{2}{c|}{128} &   & \multicolumn{2}{c|}{64} & \multicolumn{2}{c|}{128} &  \\  
        Sequence & 00  & 22 & 25 &  55 &  Avg. & 00  & 22 & 25 &  55 &  Avg.\\  \hline
        Cartographer   &    -/-    &    -/- &      -/- & -/-    &      -/-  &    -/-    &    -/- &      -/- & -/-    &      -/-    \\ 
        
        HDL-Graph & 5.11/19.9 &      3.16/12.3   & 3.42/36.8 & 7.69/22.1  & 4.85/22.8  & \textbf{0.34}/\textbf{0.36} &      \textbf{0.30}/\textbf{0.35}   & \textbf{0.27}/\textbf{0.43} & \textbf{0.30}/\textbf{0.47}  & \textbf{0.30}/\textbf{0.40}   \\ 
        
        SC-LL &     4.03/8.75   & 3.48/14.8 & \textbf{2.76}/25.1  &  5.06/\textbf{10.92} & 3.83/14.9 &     0.81/0.64   & 0.33/0.51 &  0.53/0.93  &  0.77/1.02 & 0.61/0.77\\ 
        
        SC-LIO SAM &        \textbf{2.52/6.05}    & \textbf{2.61}/\textbf{11.2} & 3.73/\textbf{18.2} & \textbf{3.32}/11.0 & \textbf{3.04}/\textbf{11.6} &        0.35/0.53    &   \textbf{0.30}/0.54 & 0.51/1.03 & 0.54/0.86 & 0.42/0.74   \\
        \noalign{\hrule height 1pt}\hline	
    \end{tabular}
    }
    \label{tab:fuberlin_results}
\end{table*}

\begin{figure*}[t]
    \centering
    \begin{subfigure}[b]{0.49\textwidth}
        \includegraphics[width=\columnwidth,trim={0.5cm 1.4cm 0cm 0cm},clip]{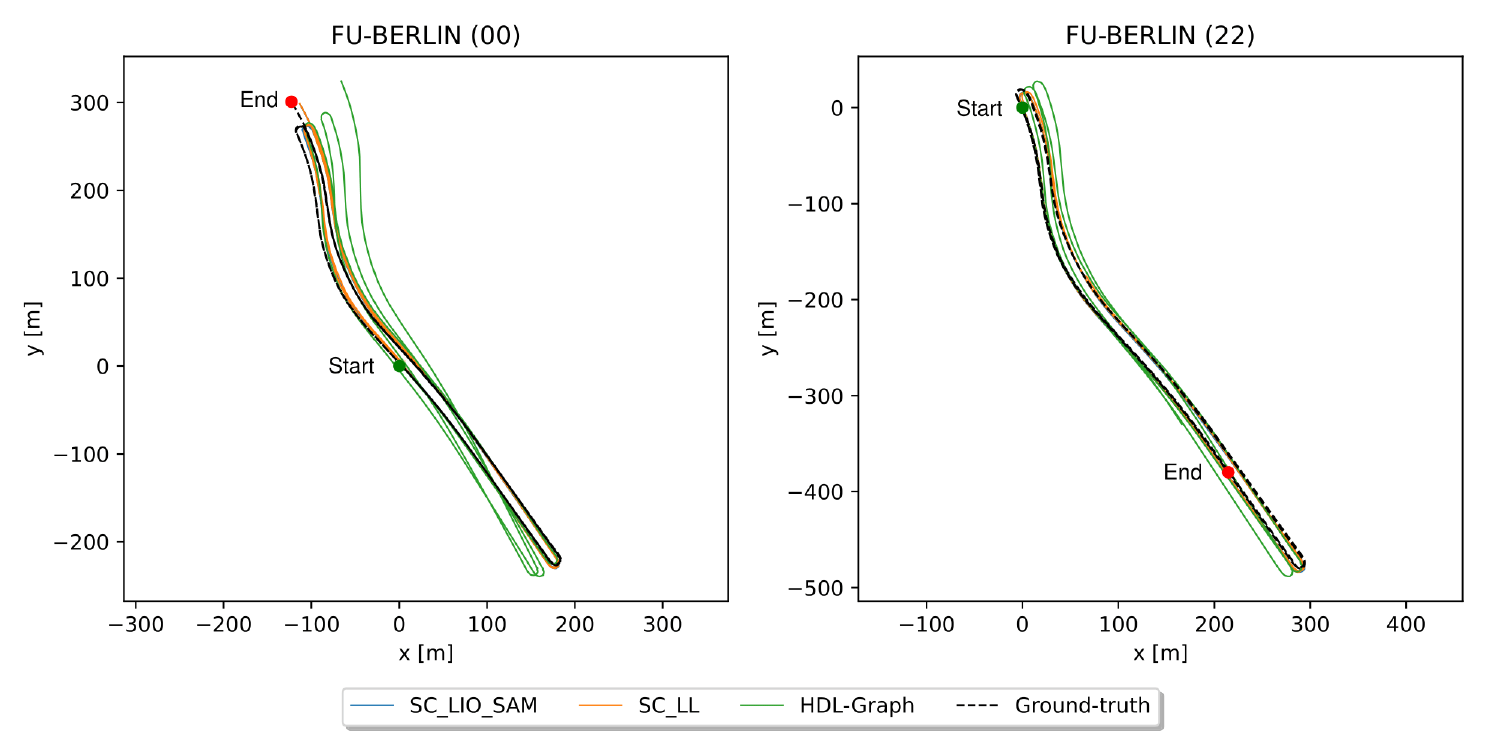}
     \end{subfigure}
    \begin{subfigure}[b]{0.49\textwidth}
        \includegraphics[width=\columnwidth,trim={0.5cm 1.4cm 0cm 0cm},clip]{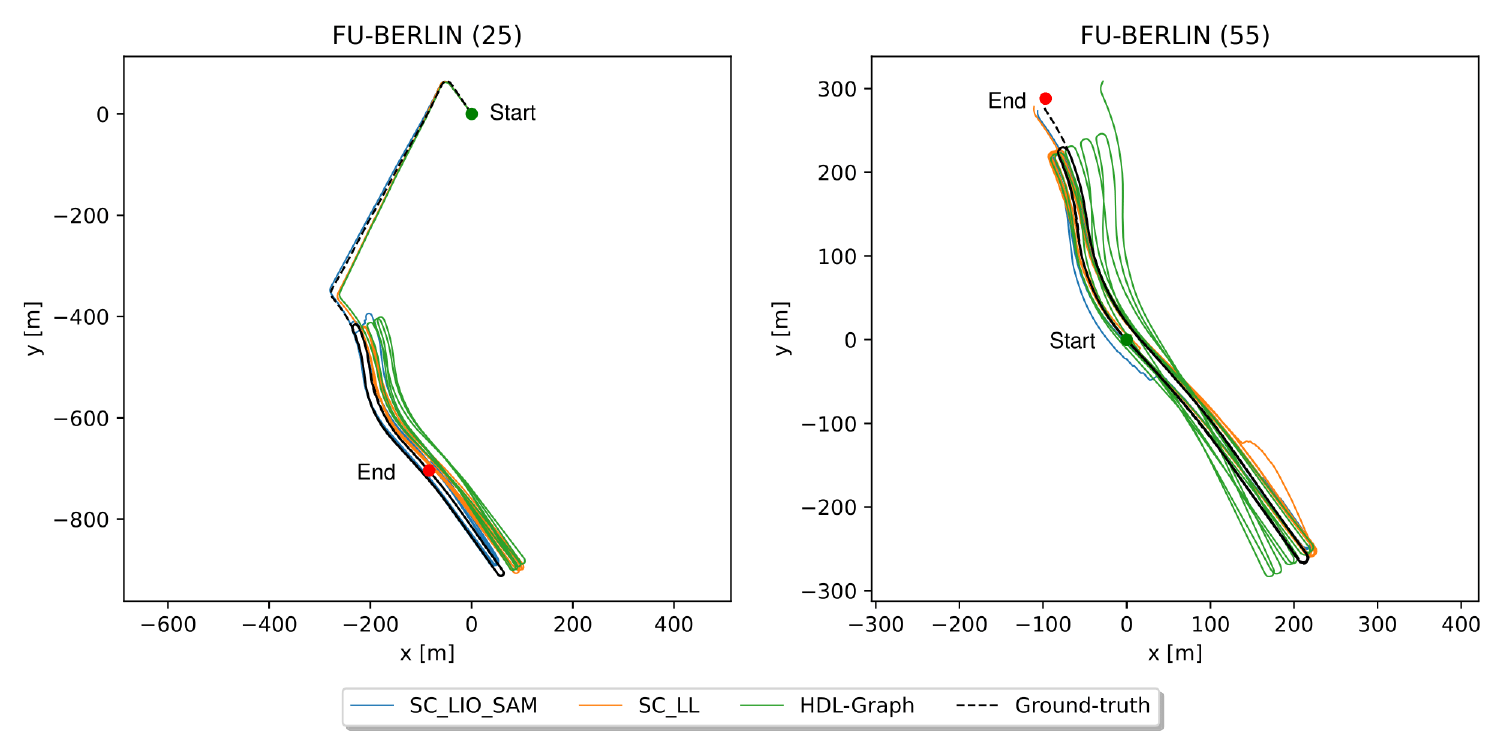}
     \end{subfigure}
     ~
     \begin{subfigure}[b]{0.49\textwidth}
        \includegraphics[width=\columnwidth,trim={0cm 0cm 0cm 0.3cm},clip]{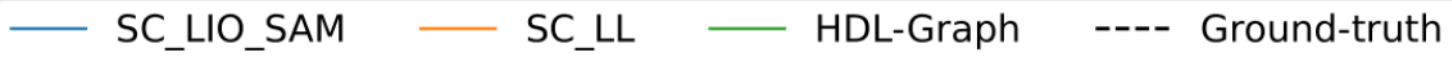}
     \end{subfigure}
    \caption{Estimated paths on the AUTONOMOS-LABS dataset.}
    \label{fig:fuberlin_results}
\end{figure*}

\subsubsection{AUTONOMOS-LABS Dataset}

Figure~\ref{fig:fuberlin_results} illustrates sequences with multiple revisits, while Table~\ref{tab:fuberlin_results} reports the performance results on the AUTONOMOS-LABS Dataset. In these sequences, HDL-Graph SLAM diverged the most from the ground-truth trajectory. On the other hand, SC-LeGO LOAM and SC-LIO SAM demonstrated a high loop detection rate. However, due to significant drift, in sequence 00, only the lower portion of the trajectory overlaps with the ground truth. The drift prevented the algorithm from perfectly aligning with the reference trajectory, although it managed not to deviate further from the repeated loop. Similar observations can be made for sequence 22, with slight differences: HDL-Graph SLAM exhibited improved performance compared to sequence 00, whereas SC-LeGO LOAM and SC-LIO SAM showed a decline in accuracy.

An analysis of sequence 25, shown in Fig.~\ref{fig:fuberlin_results} (obtained using a 128-beam LiDAR), reveals that HDL-Graph SLAM and SC-LeGO LOAM began to drift almost immediately along the straight segment. The absence of distinct features or revisited locations along this long segment indicates that the algorithms could not compensate for the accumulated error during the registration process, leading to a gradual increase in drift. In this context, graph optimization proved ineffective. The key distinction between SC-LeGO LOAM and SC-LIO SAM lies in point cloud registration. SC-LIO SAM utilizes point cloud deskewing during the registration process, a technique that is particularly advantageous when working with straight segments. Moreover, the existence of a place recognition module in SC-LIO SAM and SC-LeGO LOAM contributed to better performance in both sequences. In sequence 55, HDL-Graph SLAM exhibited significant error accumulation, failing to overlap the sequence consistently and continuing to drift away from the reference trajectory. The average ATE for SC-LIO SAM was lower, demonstrating a more accurate performance compared to SC-LeGO LOAM. Despite HDL-Graph SLAM achieving the lowest RTE, it was unable to outperform the other methods due to the lack of a robust loop closure detection algorithm.

The scan matching accuracy of HDL-Graph SLAM was superior across all sequences compared to the other methods, similar to what was observed in the KITTI dataset. This result can be attributed to the higher number of points utilized in the registration process by HDL-Graph SLAM (using FAST-GICP), compared to the feature-based approaches (edge and corner features) used by the other methods.

\subsubsection{Scan Resolution and Registration Time}
\begin{table}[t]
    \centering
    \caption{Registration time [ms] comparison of 64 and 128 LiDAR channels.  }
    {\renewcommand{\arraystretch}{1.2}
    \begin{tabular}{l|c|c}
        \noalign{\hrule height 0.5pt}\hline	
         & 64-channels & 128-channels \\ \hline 
        SC-LIO & 11.4 & 22.9  \\ 
        SC-LeGO & 14.5 & 29.1  \\ 
        HDL-Graph & 74.9  & 102.2  \\
        \noalign{\hrule height 0.5pt}\hline	
    \end{tabular}
    }
    \label{tab:registrstion_time_comparison}
\end{table}

The registration processes among SC-LIO SAM, SC-LeGO LOAM, and HDL-Graph SLAM differ. Unlike HDL-Graph SLAM, which performs registration directly on raw dense point clouds, SC-LIO SAM and SC-LeGO LOAM utilize spherical image projection followed by feature extraction and association between scans. Since the image projection and feature association modules run in separate threads, the registration time for these algorithms is defined by the time required for data extraction and association. The results are presented in Table~\ref{tab:registrstion_time_comparison}. As previously mentioned, HDL-Graph SLAM demonstrated superior performance across all sequences, albeit at the cost of increased computational time due to the larger number of points used in registration. Another notable observation is the increase in registration time from the 64-channel LiDAR to the 128-channel LiDAR. For SC-LIO SAM and SC-LeGO LOAM, this increase is approximately twofold, which aligns with the expectation that doubling the resolution would lead to a corresponding increase in matching time. This increased registration time implies a higher computational burden, which may limit real-time applicability in scenarios with limited processing power or where rapid response is critical. HDL-Graph SLAM's registration time for the 128-channel LiDAR was around $100$,ms.

\subsubsection{Map Density}

\begin{table}[t]
    \centering
    \caption{Comparison of LiDAR Map Cloud density using 64 and 128 beam LiDAR, AUTONOMOS-LABS sequences (22) and (55), respectively.}
    {\renewcommand{\arraystretch}{1.2}
    \begin{tabular}{l|ccc}
        \noalign{\hrule height 0.5pt}\hline	
        Parameter       & 64 LiDAR Map Cloud  &  128 LiDAR Map Cloud \\ \hline
        Sequence duration &   431s  &     920s   \\
        Number of Points & $\approx$59M             & $\approx$384M             \\
        Density (points/m³)        & 2.33           &  26.19               \\
        Cloud Size (GB) & 2.4                 & 13.6                \\
       \noalign{\hrule height 0.5pt}\hline	
    \end{tabular}
    }
    \label{tab:map_comparison}
\end{table}


Table~\ref{tab:map_comparison} indicates that the 128-channel LiDAR map contains a significantly higher number of points and greater density compared to the 64-channel LiDAR map. The increased density and data content of the 128-channel LiDAR map provide several benefits, such as more detailed and accurate representation of the environment, which enhances the reliability of map generation and improves feature detection. The higher resolution allows for better identification of small or distant objects, which is particularly advantageous in complex environments. A higher point density provides more detailed environmental information, which can enhance accuracy in map generation and feature detection. However, as the dataset size grows, graph optimization and loop closure—which are critical for maintaining map consistency—result in longer processing times and increased computational demands.

To address these challenges, one approach is to downsample the point clouds, thereby reducing the computational load while retaining essential information for SLAM. However, this process must be carefully managed to avoid losing important details that could compromise the overall accuracy of the map.

\section{Conclusion}
\label{sec:conclusion}

In summary, the reported experimental results on real-world datasets indicate that the SC-LIO SAM and Cartographer achieve quite good results overall. Also, based on the quantitative and qualitative comparisons carried out,  SC-LIO SAM and Cartographer achieved notable performance in all sequences. Both SC-LeGO LOAM and SC-LIO SAM were capable of detecting revisited places when necessary, thereby preventing the system from drifting. HDL-Graph has previously stated, was the one with the best scan matching performance, mainly due to operating on dense raw point clouds, but the absence of a robust place recognition module turned out to be the key for the error not being compensated during graph optimization. Cartographer did not have results in the FU-Berlin dataset and KITTI sequence 00 because of the missing IMU data source, however Cartographer outperformed SC-LIO by a reduced margin in ATE. The global descriptor, known as the Scan Context, is a key component of both SC-LeGO LOAM and SC-LIO SAM, as it was highly effective in terms of place recognition, which was crucial for these methods to identify the correct path as soon as they deviate from the ground-truth.


\section*{Acknowledgment}
This work has been supported by the project GreenBotics (ref. PTDC/EEI-ROB/2459/2021), founded by Fundação para a Ciência e a Tecnologia (FCT), Portugal. It was also partially supported by FCT through grant UIDB/00048/2020. T.Barros has been supported by the PhD grant with reference 2021.06492.BD.  This work has also been supported by the DFG Individual Research Grant on “Fairness and Efficiency in Emerging Vehicle Routing Problems” (497791398).


\bibliographystyle{plain}
\bibliography{refs}

\end{document}